\documentclass[10pt,twocolumn,letterpaper]{article}

\usepackage{iccv}              % To produce the CAMERA-READY version
\definecolor{iccvblue}{rgb}{0.21,0.49,0.74}
\usepackage[pagebackref,breaklinks,colorlinks,allcolors=iccvblue]{hyperref}
\usepackage{graphicx}
\usepackage{multirow}
\usepackage{latexsym}
\usepackage{amsmath}
\usepackage{mathtools}
\usepackage{amssymb}% http://ctan.org/pkg/amssymb
\usepackage{pifont}% http://ctan.org/pkg/pifont
\usepackage{colortbl} 
\usepackage[dvipsnames]{xcolor}
\newcommand{\cmark}{\ding{51}}%
\newcommand{\xmark}{\ding{55}}%

\def\paperID{601} % *** Enter the Paper ID here
\def\confName{ICCV}
\def\confYear{2025}

\newcommand{\model}{VideoPlan}

\title{Enhancing Visual Planning with Auxiliary Tasks and Multi-token Prediction}

\author{Ce Zhang$^1$\thanks{Work done during an internship at Meta.}
\quad
Yale Song$^2$
\quad
Ruta Desai$^2$
\quad
Michael Louis Iuzzolino$^2$
\\
Joseph Tighe$^2$
\quad
Gedas Bertasius$^1$
\quad
Satwik Kottur$^2$
\\
$^1$UNC Chapel Hill \quad\quad $^2$Meta
}

\begin{document}
\maketitle
\begin{abstract}
Visual Planning for Assistance (VPA) aims to predict a sequence of user actions required to achieve a specified goal based on a video showing the user’s progress. 
Although recent advances in multimodal large language models (MLLMs) have shown promising results in video understanding, long-horizon visual planning remains a challenging problem.
We identify two challenges in training large MLLMs for video-based planning tasks:
(1) scarcity of procedural annotations, limiting the model’s ability to learn procedural task dynamics effectively, and 
(2) inefficiency of next-token prediction objective to explicitly capture the structured action space for visual planning when compared to free-form, natural language.
To tackle data scarcity, we introduce Auxiliary Task Augmentation.
We design and train our model on auxiliary tasks relevant to long-horizon video-based planning (\eg, goal prediction) to augment the model’s planning ability. 
To more explicitly model the structured action space unique to visual planning tasks, we leverage Multi-token Prediction, extending traditional next-token prediction by using multiple heads to predict multiple future tokens during training.
Our approach, VideoPlan, achieves state-of-the-art VPA performance on the COIN and CrossTask datasets, surpassing prior methods by \textbf{7.3\%} and \textbf{3.4\%}, respectively, when predicting 3 future actions. 
We further extend our method to the challenging Ego4D Long-term Action Anticipation task, and show that it is on par with the state-of-the-art approaches despite not using specialized egocentric features.
Code will be made available.
\end{abstract}    
\begin{figure}[t]
    \centering
    \includegraphics[width=\columnwidth,trim=0cm 0pt 0cm 0cm, clip]{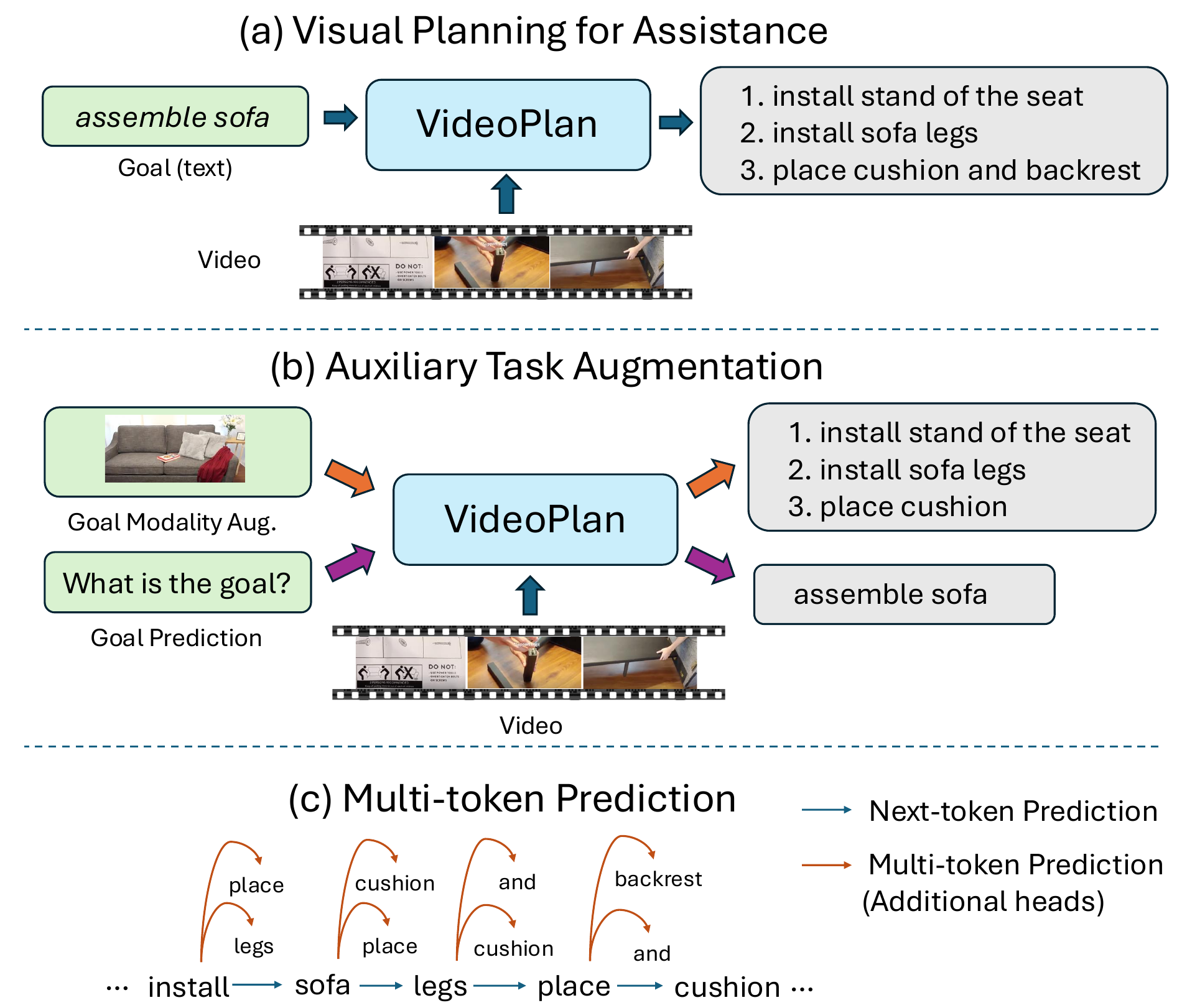}
    \caption{\textbf{(a) Visual Planning for Assistance (VPA):} predict a sequence of future actions (grey) given a video observation of user's progress and a succinct goal in text (green).
    (b) \textbf{Auxiliary Task Augmentation:} Construct additional tasks related to long-term visual planning. Inputs (green) and outputs (grey) for a given auxiliary tasks are connected via same colored arrows.
    (c) \textbf{Multi-token Prediction:} Extend next-token prediction by also modeling
    future tokens (red arrows) via additional heads. 
    VideoPlan leverages (b) and (c) to overcome data scarcity and inefficiency of next-token prediction to explicitly reason about future tokens at current step, to achieve state-of-the-art on VPA.}
    \label{fig:teaser}
    \vspace{-5mm}
\end{figure}

\section{Introduction}
\label{sec:intro}
With the rapidly increasing interest in assistive technologies, ranging from personal virtual assistant to a physical robot, the ability to anticipate future actions in a goal-oriented
 setting is crucial for such embodiments to better aid humans.
Aimed at this desired ability, Visual Planning for Assistant (VPA) task~\cite{patel2023pretrained} focuses on predicting a sequence of future actions (\eg, \textit{`install sofa legs'}, \textit{`put on sofa cover'}) necessary to achieve a specified goal (\eg, \textit{assemble sofa}), based on a video that captures the user’s progress. 
VPA has broad applications, such as helping people learn new skills (\eg, drawing, cooking) or guiding them in unfamiliar household tasks (\eg, assembly). 
To be successful in this task, a model would need to handle long context videos from untrimmed video, understand complex real-world goals, and generate consistent action plans towards them as illustrated in Fig.~\ref{fig:teaser}.

Recent developments in Multimodal Large Language Models (MLLMs) have made significant progress in video understanding, such as visual question answering~\cite{li2024llava, song2024moviechat, zhang2023video, lin2023video}, temporal grounding~\cite{huang2024vtimellm, ren2024timechat}, and visual captioning~\cite{li2023videochat}. 
Prior works~\cite{liu2023llm,hao2023reasoning,song2023llm} have shown that large language models (LLMs), which usually form the heart of MLLMs, possess procedural knowledge and can plan well in the text domain.
Inspired by this success, we investigate MLLMs as a natural modeling choice for visual planning.
However, applying MLLMs to long-horizon planning tasks like VPA uncovers two main challenges. 
First, training an MLLM-based agent capable of assisting humans with complex, long-horizon daily tasks necessitates a substantial volume of training data, each containing long sequences of procedural annotations.
Unlike short video-text pairs, procedural annotations require detailed and step-by-step labeling making data collection expensive, time consuming, and cumbersome; thereby are prohibitively resource intensive. 
The scarcity of such annotated procedural data limits the model's ability to learn the task-specific dynamics necessary for accurate action prediction.
Second, the action space for visual planning exhibits more structure compared to natural, free-form language--both in terms of individual and sequence of labels.
For instance, in a cooking scenario, the actions are limited to a set of cooking-related steps, where actions like \textit{`install sofa legs'} will never appear.
At the sequence level, there are strong long-term temporal dependencies amongst the constituent actions, \eg, \textit{`open microwave door'} is likely followed by \textit{`take an item out'} or \textit{`put an item in.'}
Traditional MLLMs are trained using the standard next-token prediction loss that shows strong generalizability when trained on a large scale of data.
However, next-token prediction might not fully capture these strong temporal structured dependencies in visual planning tasks, due to the lack of explicit reasoning
about future tokens when predicting the next token.
As a consequence of this design, the resultant models miss out on a crucial learning signal for reliable long-horizon planning, the effects of which are more pronounced in data-scarce tasks like visual planning.

In this work, we propose two strategies to address the above related challenges.
First, we introduce Auxiliary Task Augmentation, which enhances the model’s planning capabilities by training it on auxiliary tasks relevant to long-horizon visual planning. 
Specifically, we employ two types of auxiliary tasks:
(1) Goal Modality Augmentation: We modify the goal modalities, for example, changing it from text to image.
(2) Goal Prediction: The model predicts the human’s goal based on video or text inputs.
Our augmentation strategy generates additional training data, helping to address the scarcity of procedural annotations.
This ultimately enables the model to better capture human intentions and task dynamics, which are essential for effective assistance and long-horizon planning.
Second, we employ Multi-Token Prediction (MTP)~\cite{gloeckle2024better}. 
Unlike next-token prediction, which focuses on predicting only the next token, MTP introduces multiple additional heads on top of a shared model trunk during training to predict multiple future tokens simultaneously.
As a result, the model, when predicting each token, not only reasons about the next token but also explicitly reasons about the future tokens, thus essentially emulating a mild form of `planning' even at the token level.
During inference, the model removes the additional heads and generates the next token autoregressively.
In essence, MTP serves as an additional regularizer well-suited for visual planning, without sacrificing the expressibility of a language model to generate open-vocabulary actions, unlike competing approaches that use a closed-vocabulary action classifier on top~\cite{patel2023pretrained, chang2020procedure}.
Our experiments demonstrate that Multi-token Prediction captures the structured action space of planning tasks more effectively than the standard next token prediction approach, improving the model’s ability to handle the structured long-term temporal dependencies.

We conduct extensive evaluations on the COIN~\cite{tang2019coin} and CrossTask~\cite{zhukov2019cross} datasets for the VPA task. 
Both Auxiliary Task Augmentation and Multi-token Prediction individually enhance our baseline MLLM model.
Combining them, our model achieves state-of-the-art results on both datasets, outperforming previous methods~\cite{islam2024propose, patel2023pretrained, huang2022language} by $7.3\%$ (absolute) and $3.4\%$ (absolute), respectively, on success rate for predicting $3$ future steps. It is worth noting that \model~uses a smaller LLM compared with the prior SOTA method~\cite{islam2024propose}, which is important for practical applications of MLLMs for VPA in the real world~\cite{verghese2024user}.
We further extend our method to the challenging Ego4D Long-term Action Anticipation task~\cite{grauman2022ego4d}, which requires predicting 20 future actions without a specified goal.
Despite not being pre-trained on egocentric data, our approach achieves comparable results to the state-of-the-art methods.

\begin{figure*}[t]
    \centering
    \includegraphics[width=\textwidth]{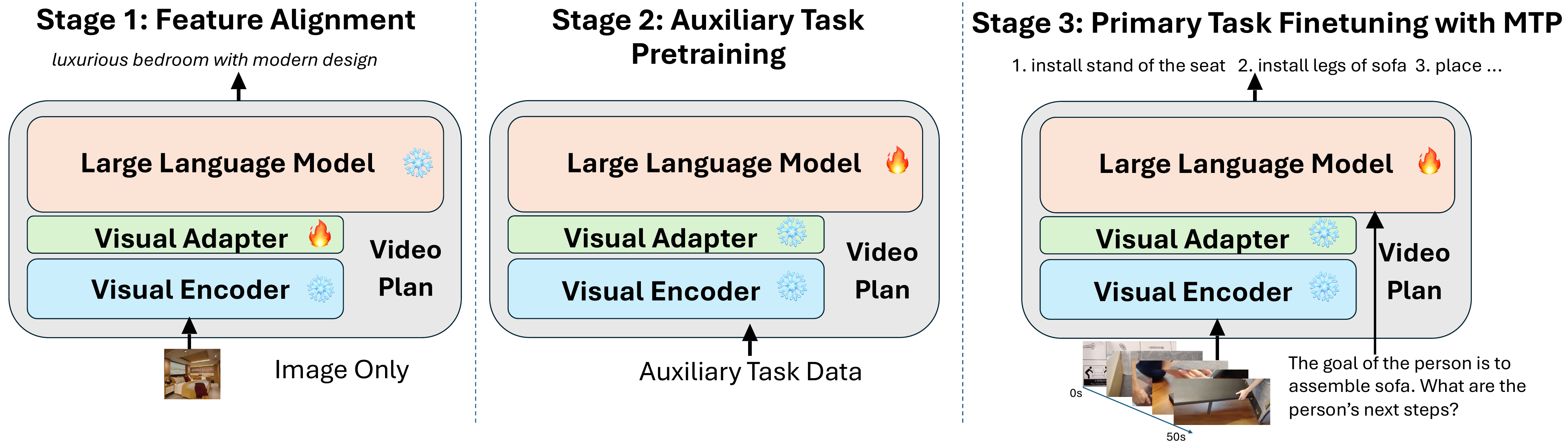}
    \caption{\textbf{Our three-stage training pipeline.} Stage 1 aligns the features of the visual encoder with the LLM embedding space by only training a visual adapter. Stage 2 helps the model better learn visual planning dynamics by training on other related auxiliary tasks.
    Finally, Stage 3 finetunes the model on VPA, the desired task at hand.
    }
    \label{fig:approach}
    \vspace{-5mm}
\end{figure*}

\section{Related Works}
\label{sec:related_work}
\textbf{Multimodal Large Language Models (MLLMs).}
Recent advancements in LLMs~\cite{touvron2023llama, achiam2023gpt, zheng2023judging} have sparked interest in MLLMs that leverage the knowledge within LLMs to enhance multimodal perception.
Flamingo~\cite{alayrac2022flamingo} integrates cross-attention modules into the LLM to handle interleaved multimodal sequences. BLIP-2~\cite{li2023blip} utilizes Q-former to encode visual information and align visual features with the LLM's input space, while LLaVA~\cite{liu2024visual} and MiniGPT-4~\cite{zhu2023minigpt} employ linear layers to connect the visual encoder and the LLM, streamlining the integration process.
To extend MLLMs to videos, most existing approaches uniformly sample frames from videos and align them with the LLM~\cite{li2023videochat, li2024llava, lin2023video}. Recent works focus on advanced frame selection or token compression methods~\cite{li2025llama, song2024moviechat, zhang2023video} to boost performance.
Recent works also explore MLLMs for downstream tasks, such as temporal grounding~\cite{huang2024vtimellm, ren2024timechat}, spatial reasoning~\cite{chen2024spatialvlm}, and planning~\cite{mu2024embodiedgpt, yang2023octopus, zheng2024towards}. For planning tasks, these works typically assume the agent can interact with the environment. Instead, EgoPlanBench~\cite{chen2023egoplan} and VPA~\cite{patel2023pretrained} introduce benchmarks and models for video-based planning without interaction, focusing on future action prediction. 
Our work also aims to enhance MLLMs' capabilities for long-horizon visual planning where interaction with the environment is infeasible.

\noindent \textbf{Planning in Instructional Videos.}
Procedure Planning~\cite{chang2020procedure} in instructional videos aims to predict multiple steps to finish the given goal based on the current observation, where both the observation and the goal are images.
Prior works employ many techniques to solve procedure planning, including using intermediate states as additional supervision~\cite{wang2023event, zhao2022p3iv, niu2024schema}, leveraging LLMs or diffusion models for planning~\cite{liu2023language, islam2024propose, wang2023pdpp, nagasinghe2024not, zare2024rap}, or using temporal prior or task prior~\cite{zhao2022p3iv, nagasinghe2024not, wang2023event, li2023skip}. However, in practice, the visual state of terminal state (goal) is not available, reducing the usefulness of procedure planning as a real-world application. To address this issue, recent work~\cite{patel2023pretrained} introduces Visual Planning for Assistance (VPA), which requires the model to output future steps based on a textual goal and a video showing the user’s progress. Our work focuses on VPA as it is a more realistic real-world assistance setting. 

\noindent \textbf{Long-term Action Anticipation (LTA).} 
This line of work aims to predict multiple future actions directly from the input video~\cite{grauman2022ego4d, nagarajan2020ego, abu2018will}. Most prior methods explore learning useful representations for future prediction~\cite{nawhal2022rethinking, ashutosh2023hiervl, zhang2024object, mascaro2023intention}. Recent works focus more on using LLMs and VLMs for LTA, leveraging the knowledge from large-scale pre-training data~\cite{mittal2024can, kim2025palm,zhao2023antgpt,wang2023vamos}. 
LTA can be viewed as a special case of long-horizon visual planning where the goal is not specified. Therefore, we evaluate our method on LTA to show the generalizability of our proposed approach.

\begin{table*}[h]
    \begin{center}
        \scalebox{1.0}{%
            \begin{tabular}{
                p{50pt}p{70pt}p{110pt}p{210pt}
            }
            \toprule[\heavyrulewidth]
            \textbf{Task Type}
            & \textbf{Observation}
            & \textbf{Instruction}
            & \textbf{Output} \\
            \midrule
            
            \multirow{3}{*}{
              \parbox{40pt}{%
                \centering
                  Goal\\Modality\\Aug.
                }%
            }
                & \includegraphics[width=50pt]{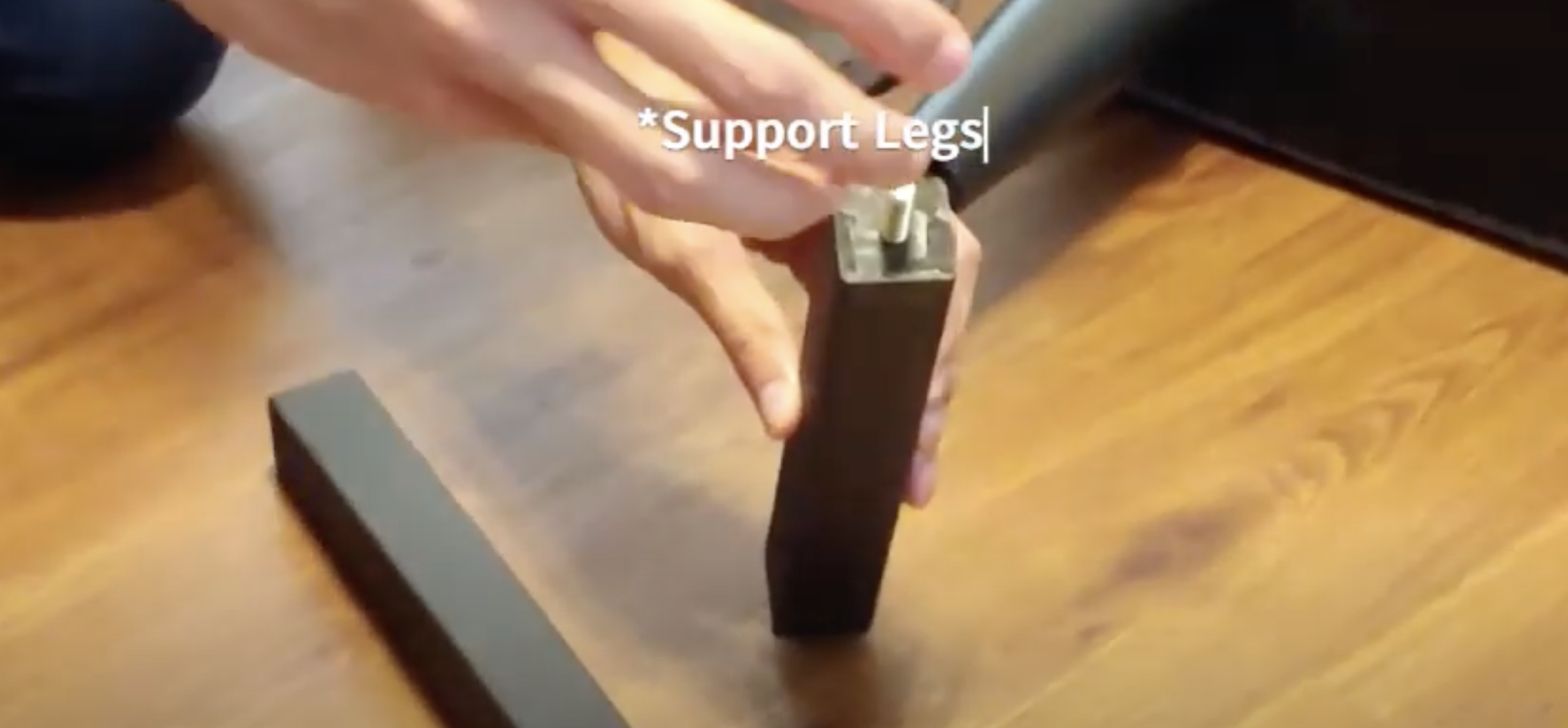} \newline
                & \parbox{130pt}{
                {\color{orange} Goal: Install Sofa.} \newline What are the next 3 steps? } 
                & \parbox{210pt}{1. install legs of sofa 2. install armrest on sofa 3. put on sofa cover.} \\

            \cmidrule(lr){2-4}
                & The legs of the sofa are separate from the base.
                & \parbox{130pt}{
                {\color{orange} Goal: \includegraphics[width=50pt]{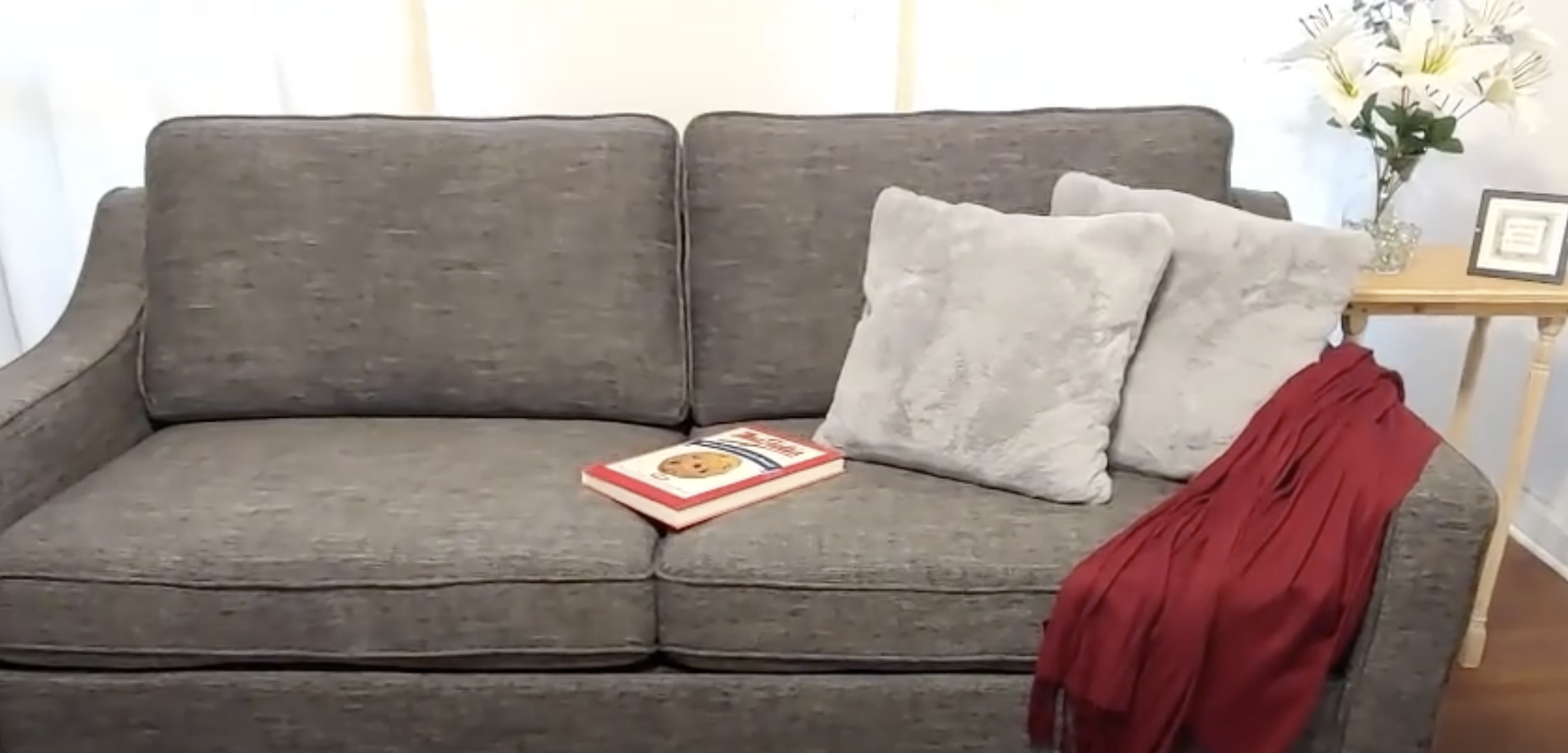}}. \newline What are the future 4 steps?} 
                & \parbox{210pt}{1. install legs of sofa 2. install armrest on sofa 3. put on sofa cover 4.put every parts mentioned together.} \\
                
            \cmidrule(lr){2-4}
                & \includegraphics[width=65pt]{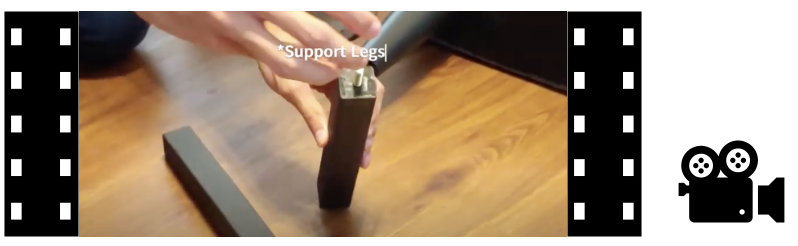}
                & \parbox{130pt}{
                {\color{orange} Goal: N/A.} \newline 
                What are the future 3 steps? } 
                & \parbox{210pt}{1. install legs of sofa 2. install armrest on sofa 3. put on sofa cover.} \\
                
            \midrule
              \parbox{40pt}{%
                \centering
                  Goal\\Prediction
                }%
                %& User's Progress (video/image/text)
                & \includegraphics[width=65pt]{figures/aux_tasks/sofa_start_video.png}
                & \parbox{130pt}{
                What is the person's goal?} 
                & \parbox{210pt}{Install Sofa}  \\
            \bottomrule[\heavyrulewidth]
            \end{tabular}
        }
    \end{center}
    \vspace*{-10pt}
    % \vspace*{\captionReduceTop}
     \caption{\textbf{List of auxiliary tasks.} For each of these, the observation can either be a video, image, or in text.
        \textbf{Goal Modality Augmentation (GMA)} varies how the {\color{orange}goal} is specified as an input.
        \textbf{Goal Prediction (GP)} requires the agent to understand the observation and predict the goal of the user. 
        }
    \label{tab:auxiliary-tasks}
    \vspace{-5mm}
\end{table*}

\section{Proposed Approach}
\label{sec:method}

In this section, we introduce \model, an MLLM designed for video-based long-horizon planning. \model~is composed of a visual encoder, a visual adapter, and an LLM (Sec.~\ref{sec:method_arch}).
To enhance planning capabilities, we adopt two key strategies: Auxiliary Task Augmentation (Sec.~\ref{sec:method_mod}) and Multi-Token Prediction (Sec.~\ref{sec:method_mtp}) that effectively tackle the shortcomings in 
naively training an MLLM for visual planning.
Finally, we describe our three stage process used to train \model~in Sec.~\ref{sec:method_stage}.

\subsection{Task Formulation}
\label{sec:method_task}
The task of Visual Planning for Assistance (VPA) aims to predict a sequence of actions $\mathcal{A} = \left \{a_1, a_2, \dots, a_H \right\}$ given the user’s goal $\mathcal{G}$ and current observation $\mathcal{O}$, where $H$ is the planning horizon~\cite{patel2023pretrained}. 
Specifically, the observation $\mathcal{O}$ is presented as an untrimmed video history capturing the user’s progress. 
The goal $\mathcal{G}$ is specified in a short natural language description (\eg \textit{assemble sofa}).
The actions $\mathcal{A}$ are annotated in free-form language (\eg \textit{`install sofa legs'}, \textit{`put on cover'}) but form a closed set of action vocabulary.

\subsection{Model Architecture}
\label{sec:method_arch}

\noindent \textbf{Visual Encoder.} 
Given a video $\mathcal{O}$, we uniformly sample $N_V$ frames represented as $V \in \mathbb{R}^{N_V \times H \times W \times C}$, where $H, W, C$ are height, width, and number of channels, respectively.
For each frame, we utilize a pretrained visual feature encoder $f_v(.)$ to extract the representation $v_i = f_v(V)$.

\noindent \textbf{Visual Adapter.} 
Following prior works~\cite{liu2024visual, moon2024anymal, patel2023pretrained}, we use a visual adapter $f_{a}(.)$ to map frame feature $v_i$ into the input embedding space of the LLM, denoted as $z_i = f_a(v_i)$.

\noindent \textbf{LLM.}
Given the goal $\mathcal{G} = \{G_i\}_{i=1}^{N_T}$, with $N_T$ number of tokens, we encode it with the model prompt using the embedding layer of the LLM $f_e(.)$, resulting in $\{g_i\}_{i=1}^{N_T}$ with $g_i = f_e(G_i)$. We then concatenate the aligned visual features $\{z_i\}_{i=1}^{T}$ with the goal text embeddings $\{g_i\}_{i=1}^{N_T}$ as the input to the LLM transformer trunk. The LLM thus generates the future action sequence conditioned on the video content and given text prompt that includes the goal text, as shown in Fig.~\ref{fig:approach}. The full text prompt is given in appendix. 

\subsection{Auxiliary Task Augmentation}

To enhance the model's planning abilities, we design a set of auxiliary tasks relevant to visual planning and generate task-specific data.
Our model is then trained jointly on the VPA and these auxiliary tasks. 
The core idea is to re-use existing annotations in novel ways that go beyond the input-output combination for VPA (see Fig.~\ref{fig:teaser}), without the need for any additional human labeling.
Different from prior instruction data construction approaches~\cite{liu2024visual, li2024mvbench}, our method is specifically designed for the visual planning task. We focus on generating instruction tuning data for long horizon video-based planning and use long instructional videos for this purpose. Additionally, Auxiliary Task Augmentation includes variants that change input modality (video to image or text) while most prior methods~\cite{liu2024visual, li2024mvbench} retain the input video. The proposed two types of auxiliary tasks and corresponding instructions are shown in Tab.~\ref{tab:auxiliary-tasks}. Below, we introduce the auxiliary tasks in detail.

\noindent \textbf{Goal Modality Augmentation (GMA).} 
\label{sec:method_mod}
In the VPA task, the observation $\mathcal{O}$ is a video and the goal $\mathcal{G}$ is a natural language text, as described in the Sec.~\ref{sec:method_task}. 
We generate modality-augmented auxiliary tasks by either changing the goal $\mathcal{G}$ from text to image or discarding it.
To convert the $\mathcal{G}$ from text to image, we utilize the existing action segment annotations in our datasets.
Specifically, we first identify the end time of the last action to be predicted, and use the corresponding last frame as the image for the goal $\mathcal{G}$.

\noindent \textbf{Goal Prediction (GP).}
Given the current observation $\mathcal{O}$, the model needs to predict the goal $\mathcal{G}$ of the person in the form of natural language. This task, along with the auxiliary tasks described above, has variations where the observation $\mathcal{O}$ is represented as either a video, text, or an image. To change $\mathcal{O}$ from video to text, we generate textual object states to replace the video. Specifically, we feed the action label and the high-level task goal to the LLMs and prompt for descriptions about possible object states before that action. More details can be found in the appendix. To change the observation from video to image modality, we simply take the last frame of the input video as the observation.

\subsection{Multi-Token Prediction}
\label{sec:method_mtp}

Recall that visual planning entails a structured action space both in terms of individual and sequence of actions.
The standard next-token prediction is too unconstrained to take advantage of such a setting, especially in a low-data regime.
Instead, we propose to use multi-token prediction~\cite{gloeckle2024better} as a way to force the
model to explicitly reason about future tokens while predicting the next-token, a beneficial trait for the visual planning task at hand.

\noindent
\textbf{Background.} Suppose $x_{1:T} = \left \{x_1, x_2, \dots, x_T \right\}$ are the input embeddings to the language model, where $T$ is the number of input tokens. Traditional language models use next-token prediction loss as the training objective:
\begin{equation}
    \mathcal{L}_{next} = - \sum_{t=1}^T \log P_\theta(x_{t+1}\ |\ x_{1:t})
\end{equation}
where $\theta$ are trainable parameters of the language model. 

In multi-token prediction~\cite{gloeckle2024better}, the model needs to predict $N$ future tokens at once during training. The training objective minimizes the cross-entropy loss for each future token:
\begin{equation}
    \mathcal{L}_{multi} = - \sum_{i=1}^N \sum_{t=1}^T \log P_\theta(x_{t+i}\ |\ x_{1:t})
\end{equation}
We consider the language model with multi-token prediction as one shared backbone and multiple output heads. Therefore, we can compute
\begin{equation}
    P_\theta(X_{t+i}\ |\ X_{1:t}) = \text{softmax}(h_i(f(X_{1:t})))
\end{equation}
where $f$ is the shared backbone and $h_i$ is the $i$-th output head. During inference, the model keeps only the next token prediction head, disabling other heads. It then generates next tokens autoregressively, as illustrated in Fig.~\ref{fig:ablation_mtp}(b).

\begin{figure}[t]
    \centering
    \includegraphics[width=0.85\columnwidth]{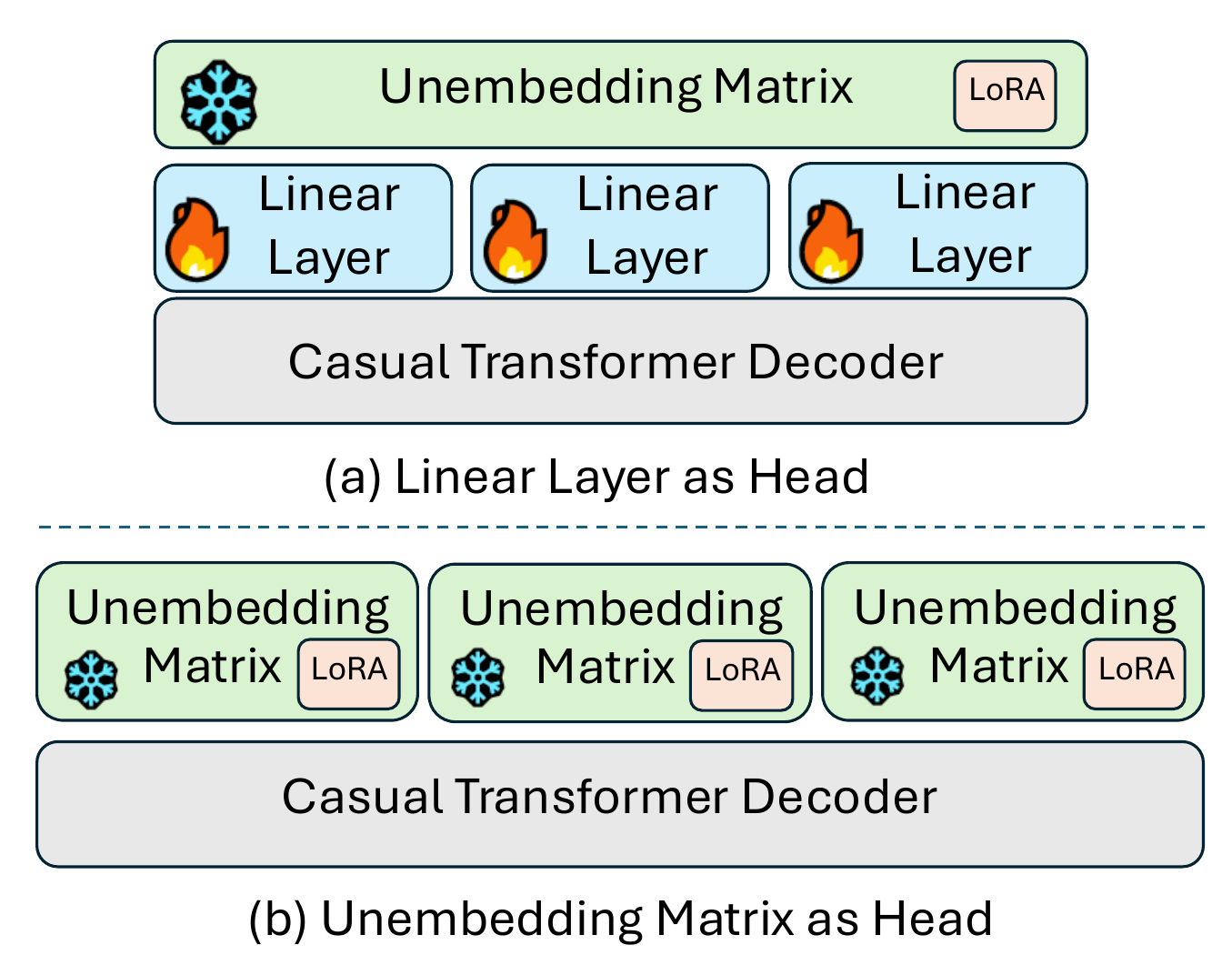}
    \vspace{-1mm}
    \caption{\textbf{Different Head Architecture for Multi-token Prediction.} \textbf{Top:} The original MTP~\cite{gloeckle2024better} introduces additional linear layers as the heads and shares the unembedding matrix on top of each head. \textbf{Bottom:} We reuse the unembedding matrix as the heads. During training, we initialize each unembedding matrix with the same pre-trained weights but add different LoRA modules.}
    \label{fig:head_arch}
    \vspace{-5mm}
\end{figure} 

Note that the original MTP~\cite{gloeckle2024better} is designed for large-scale pre-training on standard NLP tasks, while the visual planning tasks fall within the low-data regime and the output action space is more structured. Therefore, we propose a modified head architecture for MTP. As shown in Figure~\ref{fig:head_arch}, the original MTP introduces additional linear layers as the heads. All heads share the same unembedding matrix, which is used to map the output embeddings of the LLM to the token indices. As a comparison, our method does not introduce new layers by reusing the unembedding matrix. Specifically, we duplicate the pre-trained unembedding matrix for $K$ times as the heads, where $K$ is the number of future tokens to be predicted. We freeze the weights of the heads and add different LoRA modules for different heads. Section~\ref{subsec:ablations} show that our novel head architecture achieves better performance than the original MTP, despite having significantly fewer trainable parameters

\begin{table*}[!t]
\centering
\small
\setlength\tabcolsep{2pt}
\begin{tabular*}{0.9\textwidth}%
     {@{\extracolsep{\fill}}lcccccccc}
\toprule
\multicolumn{1}{l}{Method} & \multicolumn{1}{c}{Language Model} & \multicolumn{1}{c}{Visual Encoder} & \multicolumn{3}{c}{T=3}  & \multicolumn{3}{c}{T=4} \\ 
\cmidrule(lr){4-6} \cmidrule(lr){7-9}  & & & SR $\uparrow$  & mAcc $\uparrow$ & mIoU $\uparrow$  & SR $\uparrow$   & mAcc $\uparrow$   & mIoU $\uparrow$   \\ 
\midrule
DDN~\cite{chang2020procedure}  & -  & I3D  & 10.1   & 22.3  & 32.2  & 7.0   & 21.0  & 37.3          \\
LLM Baseline~\cite{touvron2023llama} & LLama-2-70B & VideoCLIP  & 10.2 & 36.6 & 50.8 & 6.1 & 30.5 & 51.5 \\
LLM Agent Baseline~\cite{huang2022language} & LLama-2-70B  & VideoCLIP & 11.1 & 40.6 & 52.8 & 6.8 & 33.5 & 53.5 \\
VLaMP~\cite{patel2023pretrained}    & GPT-2  & VideoCLIP  & 18.3   & 39.2   & 56.6 & 9.0 & 35.2  & 54.2  \\
VidAssist~\cite{islam2024propose} & LLama-2-70B  & VideoCLIP & 21.8 & 44.4 & 64.4 & 13.8 & 38.3 & 66.3 \\
\midrule
\model~(ours) & LLama-2-7B  & VideoCLIP & 25.6 & 45.3 & 67.7 & 18.2 & 43.3 & 71.9 \\
\model~(ours) & LLama-2-7B  & MetaCLIP & \textbf{29.1} & \textbf{50.1} & \textbf{69.4} & \textbf{20.5} & \textbf{47.5} & \textbf{73.9} \\
\bottomrule
\end{tabular*}
\vspace{-1mm}
\caption{\textbf{Visual Planning for Assistance on COIN.} Despite not finetuning the visual encoder, \model~achieves the best performance on all metrics. Specifically, our method outperforms the prior best-performing method, VidAssist, by \textbf{7.3\%} and \textbf{6.7\%} in Success Rate when predicting the future 3 and 4 actions, respectively. 
}
\label{tab:coin_vpa}
\vspace{-2mm}
\end{table*}

\begin{table*}[t]
\centering
\small
\setlength\tabcolsep{2pt}
\begin{tabular*}{0.9\textwidth}%
     {@{\extracolsep{\fill}}lcccccccc}
\toprule
\multicolumn{1}{l}{Method} & \multicolumn{1}{c}{Language Model} & \multicolumn{1}{c}{Visual Encoder} & \multicolumn{3}{c}{T=3}  & \multicolumn{3}{c}{T=4} \\ 
\cmidrule(lr){4-6} \cmidrule(lr){7-9}  & & & SR $\uparrow$  & mAcc $\uparrow$ & mIoU $\uparrow$  & SR $\uparrow$   & mAcc $\uparrow$   & mIoU $\uparrow$    \\ 
\midrule
DDN~\cite{chang2020procedure}  & -  & I3D    & 6.8           & 25.8          & 35.2          & 3.6          & 24.1          & 37.0          \\
LTA~\cite{grauman2022ego4d}  & -  & MViT     & 2.4           & 24.0          & 35.2          & 1.2          & 21.7          & 36.8          \\
LLM Baseline~\cite{touvron2023llama} & LLama-2-70B & VideoCLIP & 4.6 & 29.7 & 35.6 & 1.1 & 22.2 & 41.3 \\
LLM Agent Baseline~\cite{huang2022language} & LLama-2-70B  & VideoCLIP & 5.8 & 31.3 & 39.6 & 2.1 & 24.7 & 44.2 \\
VLaMP~\cite{patel2023pretrained}    & GPT-2  & VideoCLIP  & 10.3          & 35.3 & 44.0          & 4.4          & 31.7 & 43.4          \\
VidAssist~\cite{islam2024propose} & LLama-2-70B  & VideoCLIP  & 12.0 & 36.7 & 48.9 & 7.4 & 31.9 & 51.6 \\
\midrule
\model~(ours)  & LLama-2-7B & VideoCLIP & 14.4 & 37.4 & \textbf{52.0} & 8.4 & 34.9 & 53.8 \\
\model~(ours) & LLama-2-7B  & MetaCLIP & \textbf{15.4} & \textbf{39.4} & 51.4 & \textbf{9.9} & \textbf{37.4} & \textbf{54.3} \\
\bottomrule
\end{tabular*}
\vspace{-1mm}
\caption{\textbf{Visual Planning for Assistance on CrossTask.} Our method achieves the best results across all metrics without finetuning the visual encoder. Specifically, when predicting the next 3 and 4 actions, \model~outperforms the previous state-of-the-art method, VidAssist, by \textbf{3.4\%} and \textbf{2.5\%} in Success Rate, respectively. }

\label{tab:crosstask_vpa}
\vspace{-5mm}
\end{table*}

\subsection{Multi-stage Training}
\label{sec:method_stage}
Inspired by prior works~\cite{huang2024vtimellm, li2025llama}, we adopt a three-stage training pipeline to maximize the effectiveness of Auxiliary Task Augmentation and Multi-token Prediction (Fig.~\ref{fig:approach}).

\noindent \textbf{Feature Alignment.} Following common practice in training MLLMs~\cite{li2024llava, moon2024anymal, li2023blip}, we freeze both the visual encoder and the LLM, training only the visual adapter. The aligns visual features with the LLM’s input embedding space.
\newline \textbf{Auxiliary Task Pre-Training.} We freeze both the visual encoder and the visual adapter, training the LLM on all auxiliary tasks. This stage enables the model to learn task dynamics and understand user intentions, which is critical for goal-oriented planning. 
\newline \textbf{Primary Task Fine-Tuning.} Finally, we fine-tune the LLM on the VPA task directly with the visual encoder and the visual adapter frozen. In this stage, we enable Multi-token Prediction to model the structured label space. We do not use MTP in prior stages due to the different label space structure between the auxiliary tasks and the VPA task.

\subsection{Implementation Details}
Motivated by transparent data curation and privacy policies, we use MetaCLIP~\cite{dosovitskiy2020image}
with ViT-L-14~\cite{dosovitskiy2020image} as our visual encoder and Llama-2-7B~\cite{touvron2023llama} 
as our LLM, freezing the visual encoder and fine-tuning the LLM with LoRA~\cite{hu2021lora}. Following prior works~\cite{li2024llava, moon2024anymal, lin2023video}, we use a linear layer as a visual adapter.
We uniformly sample 100 frames from the input video at 0.5 FPS. When a video is shorter than 50 seconds, we sample the maximum amount of frames at 0.5 FPS. For feature alignment, we use a subset of LAION dataset with ~$550K$ image-text pairs.
We use 4 additional heads for MTP, predicting 4 future tokens in addition to the next token during training.
We provide more implementation details in the appendix. 
\section{Experiments}

\begin{table*}[t]
\centering
\small
\begin{tabular*}{0.8\textwidth}%
     {lcccccc}
\toprule
\multicolumn{1}{c}{Method} & Language Model & Visual Encoder & ED (verb) $\downarrow$ & ED (noun) $\downarrow$ &ED (action) $\downarrow$ \\ 
\midrule
ObjectPrompt~\cite{zhang2024object} & - & CLIP  & 0.7004 & 0.7092 & 0.9142 \\
PlausiVL~\cite{mittal2024can} & LLama-2-7B & CLIP  & 0.679 & 0.681 & - \\
AntGPT~\cite{zhao2023antgpt} & LLama-2-70B & CLIP  & 0.6531 & 0.6446 & 0.8748 \\
Vamos~\cite{wang2023vamos}  & LLama-2-7B & CLIP  & 0.643 & 0.650 & 0.868 \\
\rowcolor{gray!20} \textcolor{gray}{EgoVideo}~\cite{pei2024egovideo} & \textcolor{gray}{Vicuna-7B} & \textcolor{gray}{EgoVideo} & \textcolor{gray}{0.6354} & \textcolor{gray}{0.6367} & \textcolor{gray}{0.8504} \\
\rowcolor{gray!20} \textcolor{gray}{PALM}~\cite{kim2025palm} & \textcolor{gray}{LLama-2-13B} & \textcolor{gray}{EgoVLP} & \textcolor{gray}{0.6471} & \textbf{0.6117} & \textbf{0.8503} \\
\midrule
\model~(ours) & LLama-2-7B & MetaCLIP & 0.6491 & 0.6504 & 0.8746 \\
\model~(ours) & Vicuna-7B & CLIP & \textbf{0.6340} & 0.6395 & 0.8649 \\
\bottomrule
\end{tabular*}
\vspace{-1mm}
\caption{\textbf{Long-term Action Anticipation on Ego4D.} We report our model's performance on the test set, and \textcolor{gray}{de-emphasize} the methods that are pretrained on large-scale egocentric data. Our method achieves the lowest edit-distance on verb prediction and competitive results on noun and action prediction. Compared with prior methods that are not pretrained on egocentric data, our model achieves the best performance across all metrics. }
\label{tab:ego4d_lta}
\vspace{-5mm}
\end{table*}

\subsection{Setup}
We evaluate our method on two tasks: Visual Planning for Assistance (VPA) and Long-term Action Anticipation (LTA).
For VPA, we evaluate on two widely used instructional video datasets, COIN~\cite{tang2019coin} and CrossTask~\cite{zhukov2019cross}. For LTA, we evaluate on Ego4D~\cite{grauman2022ego4d}. 
Both VPA and LTA use untrimmed videos as input. VPA provides a specified goal to the model, whereas LTA focuses on action anticipation without an explicit goal. Additionally, the planning horizon in VPA is set to predict 3 or 4 steps, while LTA requires the model to anticipate 20 actions in the future.

\noindent \textbf{Datasets.} The COIN~\cite{tang2019coin} dataset is a large-scale instructional video dataset designed for understanding complex tasks across various domains. It includes over 11,827 videos with 180 different tasks, such as cooking, DIY, and other household activities. The actions are labeled with one natural language description (\eg, \textit{`install sofa leg'}), start time, end time, and the high-level task (\eg, \textit{`assemble sofa'}). On average, each video is 2.4 minutes long and includes about 3.6 labeled actions. The CrossTask~\cite{zhukov2019cross}. dataset includes 2,750 videos from 18 procedural tasks. Each video is approximately 5 minutes long on average and contains around 7.6 annotated actions. Ego4D~\cite{grauman2022ego4d} contains over 3,600 hours of egocentric video of daily life activity spanning hundreds of scenarios. We focus on the subset for long-term action anticipation, which contains 3,472 annotated clips with a total duration of around 243 hours. The actions are labeled with one verb and one noun. The dataset has 117 types of verbs and 521 types of nouns in total.

\noindent \textbf{Metrics.} For VPA, we evaluate our model on three metrics. (1) Success Rate (SR) is the strictest metric. It considers the predicted sequence of actions to be success only if every predicted action is correct. (2) mean Accuracy (mAcc) calculates the average accuracy of the predicted actions at every step. (3) Mean Intersection over Union (mIoU) treats the predicted and ground truth actions as two sets. It calculates the average Intersection over Union between the predicted action set and the ground action set across the test set. For the LTA task, we calculate the edit-distance (ED) following the Ego4D LTA setup. Specifically, we report the minimum edit distance among 5 predicted verb, noun, and action sequences, for a horizon of $20$ actions.

\subsection{Main Results}
\label{sec:main-results}
\textbf{VPA.} We compare our method with prior methods on the COIN and CrossTask datasets. The results are presented in Table~\ref{tab:coin_vpa} and Table~\ref{tab:crosstask_vpa}, respectively. We observe that \model~outperforms the previous methods by a large margin in all metrics. 
Specifically, on COIN dataset, \model~achieves $+7.4\%$ and $+5.7\%$ higher Success Rate than VidAssist~\cite{islam2024propose} for predicting $T=3$ and $T=4$ future steps. 
Similarly, on the CrossTask dataset, our method outperforms the prior state-of-the-art methods by $+3.4\%$ and $+2.5\%$ in Success Rate, respectively. 
For fair comparison with prior methods, we also include a variant of our method with VideoCLIP as the visual encoder. From Table~\ref{tab:coin_vpa} and Table~\ref{tab:crosstask_vpa} we can see that with the same visual encoder, our method still outperforms VidAssist~\cite{islam2024propose} on both COIN and CrossTask dataset across all metrics. Specifically, our method with VideoCLIP outperforms VidAssist~\cite{islam2024propose} by $+3.8\%$ and $+4.4\%$ in Success Rate when planning for the future $T=3$ and $T=4$ future steps. Similarly, on the CrossTask dataset, our method outperforms VidAssist~\cite{islam2024propose} by $+2.4\%$ and $+1.0\%$ in Success Rate, respectively. These results indicate that the strong performance of our model come from the proposed modules (i.e., ATA and MTP) rather than the visual encoder alone.
Additionally, our language model only has 7B parameters compared to prior LLM-based methods which use language models with more than 70B parameters. These results highlight the superior video-based planning ability of our proposed method. 

\noindent \textbf{LTA.} 
In Table~\ref{tab:ego4d_lta}, we compare our method with prior methods on the Ego4D LTA benchmark. 
Many existing methods~\cite{pei2024egovideo, kim2025palm} leverage large-scale egocentric pretraining data for action anticipation. In contrast, our method does not rely on such egocentric pretraining.
Despite this difference, our model still achieves the lowest edit distance on verb prediction and competitive results on noun and action prediction. Notably, when compared with prior methods that are also not pretrained on egocentric data, our model outperforms all others across all metrics. These results validate the generalizability of our proposed method.

\subsection{Ablations and Discussion}
\label{subsec:ablations}

\begin{table}[t]
    \centering
    \small
    \setlength\tabcolsep{4pt}
    \begin{tabular}{cc|cccccc}
    \toprule
         ATA & MTP & \multicolumn{3}{c}{T=3}  & \multicolumn{3}{c}{T=4} \\ 
\cmidrule(lr){3-5} \cmidrule(lr){6-8} &  & SR & mAcc  & mIoU & SR & mAcc & mIoU \\
         \midrule
         \xmark & \xmark & 25.7 & 46.4 & 67.4 & 17.5 & 43.6 & 71.9 \\
         \cmark & \xmark & 27.9 & 48.6 & 67.3 & 18.8 & 45.1 & 70.8 \\
         \xmark & \cmark & 27.7 & 48.2 & 68.7 & 19.2 & 45.1 & 73.3 \\
         \cmark & \cmark & \textbf{29.1} & \textbf{50.1} & \textbf{69.4} & \textbf{20.5} & \textbf{47.5} & \textbf{73.9} \\
    \bottomrule
    \end{tabular}
    \vspace{-1mm}
    \caption{\textbf{Effects of Auxiliary Task Augmentation and Multi-token Prediction on COIN dataset.} Both Auxiliary Task Augmentation (ATA) and Multi-token Prediction (MTP) improve planning ability. 
    Our best model (ATA + MTP) outperforms the baseline model (without ATA and MTP) by \textbf{3.4\%} and \textbf{3.0\%} on Success Rate when predicting the next 3 and 4 actions respectively.}
    \label{tab:ablation_main}
    \vspace{-5mm}
\end{table}

In this section, we perform various ablation studies on Auxiliary Task Augmentation and Multi-token Prediction. We conduct all experiments on the COIN dataset.

\noindent \textbf{Auxiliary Task Augmentation.} 
We study the effects of Auxiliary Task Augmentation (ATA) in Table~\ref{tab:ablation_main}. From the results, we can observe that ATA consistently improves the model’s performance across most of the metrics. Specifically, when predicting 3 and 4 future actions, ATA improves our baseline model (without MTP and ATA) by $2.2\%$ and $1.3\%$ in success rate, respectively. Additionally, ATA enhances the success rate of the MTP-only model by $1.4\%$ and $1.3\%$, respectively. These results show the effectiveness of ATA in enhancing the model’s visual planning ability.

\noindent \textbf{Multi-token Prediction.} Table~\ref{tab:ablation_main} shows the impact of Multi-token Prediction (MTP) on model performance. The results show that MTP consistently boosts model performance across all  metrics. Specifically, when predicting 3 and 4 future actions, MTP enhances the success rate of the our baseline model (without MTP and ATA) by $2.0\%$ and $1.7\%$, respectively. Additionally, MTP improves the ATA-only model by $1.2\%$ and $1.7\%$ in success rate, respectively. These results demonstrate the effectiveness of MTP in improving the model’s visual planning abilities.

\begin{table}[t]
    \centering
    \small
    \setlength\tabcolsep{4pt}
    \begin{tabular}{c|cccccc}
    \toprule
         Aux. Tasks & \multicolumn{3}{c}{T=3}  & \multicolumn{3}{c}{T=4} \\ 
\cmidrule(lr){2-4} \cmidrule(lr){5-7} & SR & mAcc  & mIoU & SR & mAcc & mIoU \\
         \midrule
         All & \textbf{29.1} & 50.1 & \textbf{69.4} & \textbf{20.5} & \textbf{47.5} & \textbf{73.9} \\
         \midrule
         w.o. GMA & 28.1 & 48.8 & 69.0 & 19.6 & 46.3 & 73.0 \\
         w.o. GP & 28.9 & \textbf{50.3} & 69.2 & 19.6 & 46.9 & 72.9 \\
         \midrule
         w.o. All & 27.7 & 48.2 & 68.7 & 19.2 & 45.1 & 73.3 \\
    \bottomrule
    \end{tabular}
    \vspace{-1mm}
    \caption{\textbf{Breakdown of Auxiliary Tasks on COIN dataset.} GMA: Goal Modality Augmentation. GP: Goal Prediction. Our model with all auxiliary tasks performs the best in SR for both $T=3$ and $T=4$. Removing GMA and removing GP both lead to a drop in SR, indicating that each auxiliary task improves the model’s planning ability.}
    \label{tab:ablation_aux}
    \vspace{-5mm}
\end{table}
\noindent \textbf{Auxiliary Tasks Analysis.}
We analyze the effect of each auxiliary task type by removing them from our best-performing model variant. As shown in Table~\ref{tab:ablation_aux}, our model with all auxiliary tasks performs best in SR on the COIN dataset when planning for the both T=3 and T=4 future action steps. Specifically, when removing Goal Modality Augmentation (GMA), we observe $1.0\%$ and $0.9\%$ drop in Success Rate for T=3 and T=4.
Meanwhile, removing Goal Prediction (GP) leads to $0.2\%$ and $0.9\%$ drop in Success Rate, respectively. 
These results indicate that each auxiliary task improves the model’s planning ability.
We hypothesize that GMA brings the most significant performance improvement because introducing goal specifications (text, image, or video) allows the model learn cross-modal goal dependencies via step actions, leading to better generalization, e.g., wash carrots is a good first step for a goal make carrot cake or any goal of a dish with carrots.

\begin{table}[h]
    \centering
    \small
    \setlength\tabcolsep{8pt}
    \begin{tabular}{cc|ccc}
    \toprule
         Head & Head & \multicolumn{3}{c}{T=3} \\ 
        \cmidrule(lr){3-5} Type & Params & SR & mAcc & mIoU \\
         \midrule
         Linear Layer~\cite{gloeckle2024better} & 80M & 28.2 & 49.4 & 68.5 \\
         \midrule
          Unembedding & \multirow{2}{*}{11M} & \multirow{2}{*}{\textbf{29.1}} & \multirow{2}{*}{\textbf{50.1}} & \multirow{2}{*}{\textbf{69.4}} \\
         Matrix (ours) & &  \\
    \bottomrule
    \end{tabular}
    \vspace{-1mm}
    \caption{\textbf{Performance of Different Head Architectures on COIN Dataset.} The head parameters are computed with 4 extra heads. Ours outperforms the original MTP~\cite{gloeckle2024better} across all metrics, showing the effectiveness of our head architecture.}
    \label{tab:ablation_head}
    \vspace{-3mm}
\end{table}

\noindent \textbf{Head Architecture for Multi-token Prediction.}
We compare the performance of our head design with the original MTP~\cite{gloeckle2024better} in Table~\ref{tab:ablation_head}. Our head design achieves the best results across all metrics, indicating the effectiveness of our method. 
We hypothesize this is because our head design leads to less trainable parameters (11M) compared to the original MTP~\cite{gloeckle2024better} (80M). The extra heads will be discarded during inference. We hypothesize that our lightweight head design forces the model to update the shared LLM backbone more than the extra heads, thus enabling it to better capture the structured action space. 

\begin{figure}[t]
    \centering
    \includegraphics[width=\columnwidth, trim=150pt 128pt 200pt 30pt, clip]{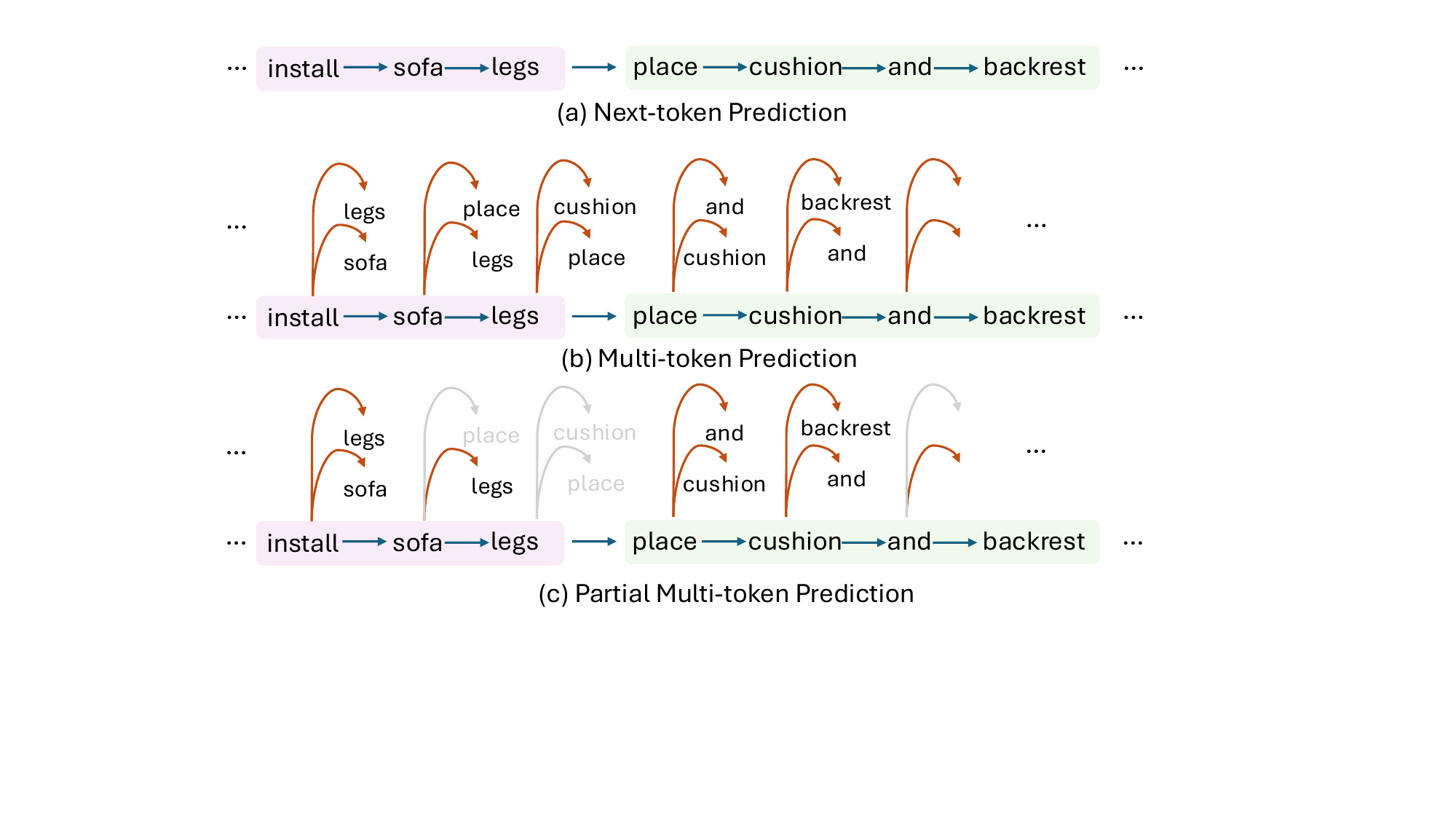}
    \caption{Token prediction schemes operating on a sequence of actions. (a) \textbf{Next-token Prediction (NTP)} only considers the following token (blue arrows). 
    (b) \textbf{Multi-token Prediction (MTP)} also reasons about future tokens via parallel token heads (red arrows). 
    (c) \textbf{Partial MTP} is an ablation on MTP where additional heads are constrained to not extend beyond action boundaries 
    (gray arrows are inactive heads).
    Our experiments show MTP outperforms both NTP and partial MTP, thus more effectively capturing the temporal dependency in the label space for visual planning.
    }
    \label{fig:ablation_mtp}
    \vspace{-5mm}
\end{figure}
\begin{table}[]
    \centering
    \small
    \setlength\tabcolsep{4pt}
    \begin{tabular}{c|cccccc}
    \toprule
         Method & \multicolumn{3}{c}{T=3}  & \multicolumn{3}{c}{T=4} \\ 
\cmidrule(lr){2-4} \cmidrule(lr){5-7} & SR & mAcc  & mIoU & SR & mAcc & mIoU \\
         \midrule
         NTP & 27.9 & 48.6 & 67.3 & 18.8 & 45.1 & 70.8 \\
         partial-MTP & 28.1 & 48.7 & 69.0 & 20.0 & 45.7 & 72.3 \\
         MTP & \textbf{29.1} & \textbf{50.1} & \textbf{69.4} & \textbf{20.5} & \textbf{47.5} & \textbf{73.9} \\
    \bottomrule
    \end{tabular}
    \caption{\textbf{Comparison between Multi-token Prediction (MTP), Next-token Prediction (NTP) and partial-MTP.} Partial-MTP allows only the standard next-token prediction to handle consequential transitions while disabling multi-token prediction. When planning for the next 3 and 4 steps, MTP outperforms partial-MTP by 3.9\% and 1.2\% in Success Rate, respectively. These results show that MTP is effective in modeling the consequential transitions in the structured label space of the planning task.}
    \label{tab:ablation_mtp}
    \vspace{-5mm}
\end{table}
\noindent \textbf{Role of MTP in goal-based visual planning.} 
Gloeckle et al.~\cite{gloeckle2024better} posit that Multi-token Prediction (MTP) assigns higher implicit weights to \textit{consequential token transitions}--transitions that are more ambiguous to predict but significantly influence the subsequent generation--thus improving the quality of overall generation.
Visual planning tasks, by virtue of the temporal structure in the target action sequences, inherits many such consequential token transitions.
Consider the sequence of actions shown in Fig.~\ref{fig:ablation_mtp}, in particular
the transition between two actions: \textit{legs}~$\rightarrow$~\textit{place}.
Predicting the leading verb \textit{place} incurs a larger ambiguity due to nature of visual planning task compared to the subsequent nouns \textit{cushion and backrest} given rest of the context.
Additionally, correctly predicting \textit{place} is more critical for the fidelity of the action in the plan compared to remainder tokens within the action, thus making it a consequential transition.
We hypothesize that MTP captures these inter-action transitions better by utilizing the structure in the label space.
To verify this, we design an ablated version of MTP where the additional heads operate only for tokens within a given action, and remain inactive across the action boundary (shown as gray arrows in Fig.\ref{fig:ablation_mtp}(c)).
This partial-MTP setting allows only the standard next-token prediction to handle consequential transitions while disabling multi-token prediction.
From Tab.~\ref{tab:ablation_mtp}, we observe that the performance of partial-MTP significantly drops compared to MTP.
Specifically, when predicting for the future 3 actions, the Success Rate drops by $1.0\%$. When predicting for the future 4 actions, the mAcc and mIoU drops by $2.2\%$ and $1.6\%$, respectively. These findings indicate that partial-MTP is less effective in modeling consequential token transitions, confirming our hypothesis that MTP plays a critical role in capturing temporal dependencies within the structured label space of our task.

\section{Conclusion}
\label{sec:conclusion}
We introduce \model, a multimodal large language model optimized for long-horizon visual planning. To tackle the data scarcity issue of procedural annotations, we introduce Auxiliary Task Augmentation (ATA). To more explicitly leverage the structured action space unique to visual planning tasks, we adapt Multi-token Prediction (MTP) for visual planning, extending traditional next-token prediction by using multiple heads to predict multiple future tokens. Extensive experiments show that both ATA and MTP improves the model's planning ability. Our approach achieves state-of-the-art results on the COIN and CrossTask datasets for the VPA task. On the Ego4D LTA task, our method achieve the SOTA results in verb prediction and competitive results on both noun and action prediction without any large-scale egocentric pretraining, demonstrating its strong generalizability.
{
    \small
    \bibliographystyle{ieeenat_fullname}
    \bibliography{main}
}

\clearpage
\setcounter{page}{1}
\maketitlesupplementary

This supplementary material is organized as follows. First we provide additional experiments in Section~\ref{sec:additional_experiments}. Then we provide more details about the training and evaluation process of our method in Section~\ref{sec:additional_implementation_details}. Finally we provide qualitative analysis of our method for both VPA and LTA tasks in Section~\ref{sec:qualitative_analysis}.

\section{Additional Experiments}
\label{sec:additional_experiments}
\subsection{Ablations on Visual Encoder}
Table~\ref{tab:ablation_visual_encoder} shows the performance of different visual encoders in our model. We use Llama2-7B as the LLM for all experiments. The results show that MetaCLIP as the visual encoder outperforms VideoCLIP across all metrics.
\begin{table}[h]
    \centering
    \small
    \setlength\tabcolsep{4pt}
    \begin{tabular}{c|cccccc}
    \toprule
         Visual & \multicolumn{3}{c}{T=3}  & \multicolumn{3}{c}{T=4} \\ 
\cmidrule(lr){2-4} \cmidrule(lr){5-7} Encoder & SR & mAcc  & mIoU & SR & mAcc & mIoU \\
         \midrule
         VideoCLIP & 25.6 & 45.3 & 67.7 & 18.2 & 43.3 & 71.9 \\
         % CLIP & \textbf{29.3} & \textbf{51.3} & \textbf{69.8} & \textbf{20.9} & \textbf{48.3} & \textbf{74.0} \\
         MetaCLIP & \textbf{29.1} & \textbf{50.1} & \textbf{69.4} & \textbf{20.5} & \textbf{47.5} & \textbf{73.9} \\
    \bottomrule
    \end{tabular}
    \caption{\textbf{Ablations on Visual Encoder on COIN Dataset.} We use Llama2-7B as the LLM. MetaCLIP as the visual encoder outperforms VideoCLIP across all metrics.}
    \label{tab:ablation_visual_encoder}
\end{table}

\begin{table}[]
    \centering
    \small
    \setlength\tabcolsep{4pt}
    \begin{tabular}{c|cccccc}
    \toprule
         Aux. Tasks & \multicolumn{3}{c}{T=3}  & \multicolumn{3}{c}{T=4} \\ 
\cmidrule(lr){2-4} \cmidrule(lr){5-7} & SR & mAcc  & mIoU & SR & mAcc & mIoU \\
         \midrule
         w. SP & \textbf{29.2} & \textbf{50.7} & \textbf{69.5} & 19.5 & 46.7 & 73.1 \\
         w.o. SP & 29.1 & 50.1 & 69.4 & \textbf{20.5} & \textbf{47.5} & \textbf{73.9} \\
    \bottomrule
    \end{tabular}
    \caption{\textbf{Effects of State Prediction.} SP: State Prediction. When planning for the next 3 future actions, our model with all auxiliary tasks performs best across all metrics. When planning for the next 4 actions, our model without SP leads to the best results. }
    \label{tab:ablation_sp}
\end{table}

\begin{table*}[t]
    \begin{center}
        \scalebox{1.0}{%
            \begin{tabular}{
                p{50pt}ccp{220pt}
            }
            \toprule[\heavyrulewidth]
            % \multirow{2}{*}{\textbf{Task Type}}
            % & \multicolumn{2}{c}{\textbf{Inputs}}
            % & \multirow{2}{*}{\textbf{Output}}
            % & \multirow{2}{*}{\textbf{Prompt}} \\
            \textbf{Task Type}
            & \textbf{Input}
            & \textbf{Output}
            & \textbf{Psuedo-Prompt} \\
            \midrule
            
            \multirow{3}{*}{
              \parbox{40pt}{%
                \centering
                  Goal\\Modality\\Aug.
                }%
            }
                %& User's Progress (video/image/text)
                & Goal (text)
                & Action (text)
                & \textit{\texttt{<obs>} The person is trying to achieve \texttt{<goal text>}. What are the next steps?} \\

            \cmidrule(lr){2-4}
                & Goal (image)
                & Action (text)
                & \textit{\texttt{<obs>} The person is trying to achieve the goal \texttt{<goal image>}. What are the next steps?} \\

            \cmidrule(lr){2-4}
                & -
                & Action (text)
                & \textit{\texttt{<obs>} What are the next steps of the person?} \\
                % & $0.895$ \\ %\reportval{0.884}{0.001} \\
            \midrule
              \parbox{40pt}{%
                \centering
                  Goal\\Prediction
                }%
                %& User's Progress (video/image/text)
                & -
                & Goal (text)
                & \textit{\texttt{<obs>} What is the person trying to achieve?} \\
            \midrule
                \parbox{40pt}{%
                \centering
                  State\\Prediction
                }%
                %& User's Progress (video/image/text)
                & Action (text)
                & State (text)
                & \textit{\texttt{<obs>} The person will take these \texttt{<actions>}. What are the states before and after these actions?} \\
            \bottomrule[\heavyrulewidth]
            \end{tabular}
        }
    \end{center}
    % \vspace*{-10pt}
     \caption{\textbf{Example Instructions and Responses for All Tasks.} The model takes in the instruction and generates the response. \texttt{<obs>} denotes the observation representation. During training, we replace \texttt{<obs>} with a video or an image, or current object states in the form of text. \texttt{<goal image>} denotes the embedding of the goal image.}
    \label{tab:prompts}
\end{table*}

\subsection{State Prediction as An Auxiliary Task}
\label{sec:sp}
In addition to Goal Modality Augmentation and Goal Prediction, we also explore State Prediction as an auxiliary task. Specifically, given current observation $\mathcal{O}$ and a sequence of future actions $\mathcal{A} = \left \{a_1, a_2, \dots, a_H \right\}$, the model needs to predict a sequence of future object states $\mathcal{S} = \left \{s_1, s_2, \dots, s_H \right\}$. Each object states $s_i$ is a short text (\eg \textit{`the sofa cover is stretched out and fitted onto the sofa'}) describing the object states after the user performs the action $a_i$.
However, we do not have the object states annotation. To generate the object states, a straightforward approach is to extract captions from the input video using pre-trained large vision-language models (VLMs). 
However, while state-of-the-art VLMs can reliably perform object recognition, recent works show that they consistently struggle to capture the objects’ physical states~\cite{newman2024pre}.
Motivated by prior works~\cite{niu2024schema, tateno2024learning, xue2024learning}, we leverage LLMs to generate language descriptions of object states based on their commonsense knowledge. Specifically, we feed the action label and the high-level task goal to the LLMs and prompt for descriptions about possible object states before that action. Following prior works~\cite{niu2024schema}, we adopt Chain-of-thought Prompting to first describe the details of action steps and then describe the object states according to the details of the steps.
The prompt is designed as: 

\texttt{First, describe the details of [action] for [goal] with one verb. Second, use 3 sentences to describe the object states before [action], avoiding using [verb].}

\noindent In this prompt, \texttt{[verb]} refers to the verb from the action name (\eg, \textit{`install'}) to increase the description diversity. To generate the object states after one action, we simply replace ``before'' with ``after'' in the above prompt.

\noindent Table~\ref{tab:ablation_sp} shows the results of adding State Prediction as an auxiliary task. We find that using State Prediction does not yield the optimal results. Specifically, when planning for the future 3 actions, the model variant without State Prediction (w.o. SP) is only $0.1\%$ lower in SR compared with the model variant with State Prediction (w. SP). When planning for the future 4 future actions, the w.o. SP model variant achieves the best results across all metrics. We leave the exploration in this direction for future work.

\section{Additional Implementation Details}
\label{sec:additional_implementation_details}

\subsection{Training}
We train our model for 1 epoch with a batch size of 1024. We set gradient accumulation step to 16 to reduce GPU memory usage. 
For all experiments, we use LoRA~\cite{hu2021lora} for efficient fine-tuning. 
The LoRA parameters are set to $r = 64$ and $alpha = 128$. 
In the auxiliary task pre-training stage, we set learning rate to $3e$-$4$, batch size to $1024$ and train the model for $1$ epoch. 
When finetuning the model on the VPA task, we set learning rate to $6e$-$4$, batch size to $512$ and optimize for $4$ epochs.

\subsection{Evaluation}
The output space of the MLLM is unconstrained. To evaluate our model, we need to map the free-form text to the discrete action indices. To achieve this, we use SentenceBERT~\cite{reimers2019sentence} to compute the text embeddings for the MLLM’s free-form output and all candidate actions in the datasets following prior works~\cite{islam2024propose, huang2022language, liu2023language}. Then we compare the text embedding of the free-form output with the text embeddings of all action candidates and choose the one with the highest cosine similarity as the target action.

\subsection{Prompt Design}

\noindent Our framework consists of multiple different tasks. We design task-specific prompts to handle all types of tasks. Table~\ref{tab:prompts} shows the template for designing the instructions and responses. The model takes the instructions as the input and generates the responses.

\noindent The instructions are constructed purely from the ground truth annotations from the datasets. The annotations include the start time, end time, and labels for each action. Specifically, when replacing \texttt{<obs>} with a video, we use a 50s video before the first action to predict. When replacing \texttt{<obs>} with an image, we use the frame right before the first action to predict. When replacing \texttt{<obs>} with text, we use the object states generated using the method described in Section~\ref{sec:sp}. \texttt{<goal image>} is replaced with the last frame of the last future action segment to predict as the goal image.

\noindent For VPA and Goal Modality Augmentation, the responses are the concatenation of future actions to predict. For Goal Prediction, we leverage the task type label (e.g. Assemble Sofa) from the dataset annotations as the responses. For State Prediction, we generate the responses using the method described in Section~\ref{sec:sp}.

\begin{figure*}[h]
    \centering
    \includegraphics[width=2\columnwidth]{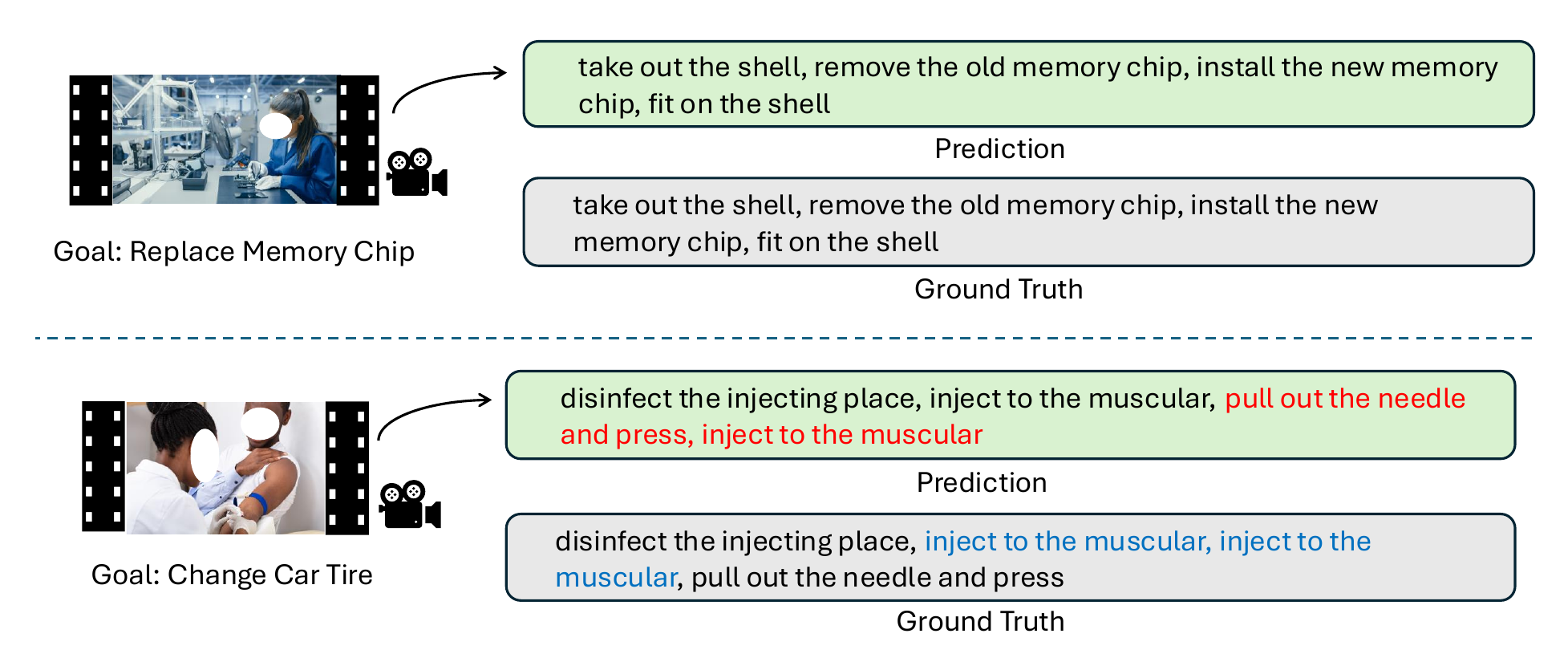}
    \caption{\textbf{Success Cases and Failure Cases from COIN Dataset.} {\color{red}The red text} denotes wrong predictions. {\color{blue}The blue text} denotes repetitive action annotations in the dataset. \textbf{Top}: One success case of our method. Our model correctly predicts all future actions. \textbf{Bottom}: One failure case of our method. The ground truth annotations contain repetitive actions ``inject to the muscular''. Our method only predicts one ``inject to the muscular''. Therefore, it begins to be incorrect from the third future action.}
    \label{fig:vis_vpa}
\end{figure*}

\begin{figure*}[h]
    \centering
    \includegraphics[width=2\columnwidth]{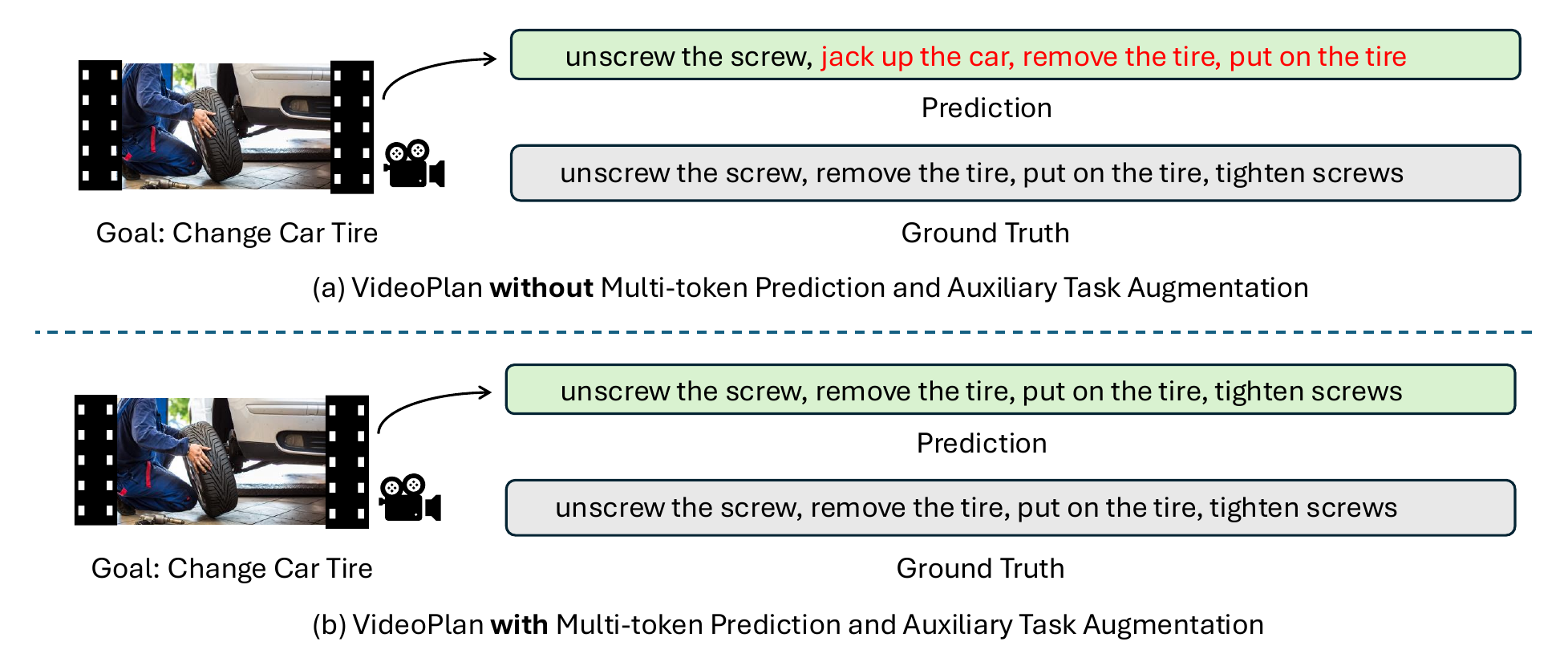}
    \caption{\textbf{Effects of Auxiliary Task Augmentation (ATA) and Multi-token Prediction (MTP).} {\color{red}The red text} denotes wrong predictions. \textbf{Top}: Our method \textbf{without} ATA and MTP. The model mistakenly outputs the action ``jack up the car'' when predicting the second action. Even though the third and the fourth predicted actions (i.e. ``remove the tire'', ``put on the tire'') match the second and the third ground truth actions, the task is still treated as failed. \textbf{Bottom}: Our method \textbf{with} ATA and MTP. Our model correctly predicts all steps.}
    \label{fig:vis_vpa_mtp_aux}
\end{figure*}

\section{Qualitative Analysis}
\label{sec:qualitative_analysis}
\textbf{VPA.} We visualize success cases and failure cases of our method in Figure~\ref{fig:vis_vpa}. The predictions in the figure are raw outputs from our method with little post-processing. Although our method generates free-from text as outputs, the raw outputs still make valid action names. From the figure we can also observe that the dataset annotations contain repetitive actions. In most cases, the repetitive actions are hard to predict. Even though our model predicts plausible actions without repetitive actions, it is still treated as failed.

\noindent In Figure~\ref{fig:vis_vpa_mtp_aux} we explore the effects of Auxiliary Task Augmentation (ATA) and Multi-token Prediction (MTP). From the figure we can observe that our method without ATA and MTP predicts the second action incorrectly while our method with ATA and MTP predict all steps correctly.

\begin{figure*}[h]
    \centering
    \includegraphics[width=2\columnwidth]{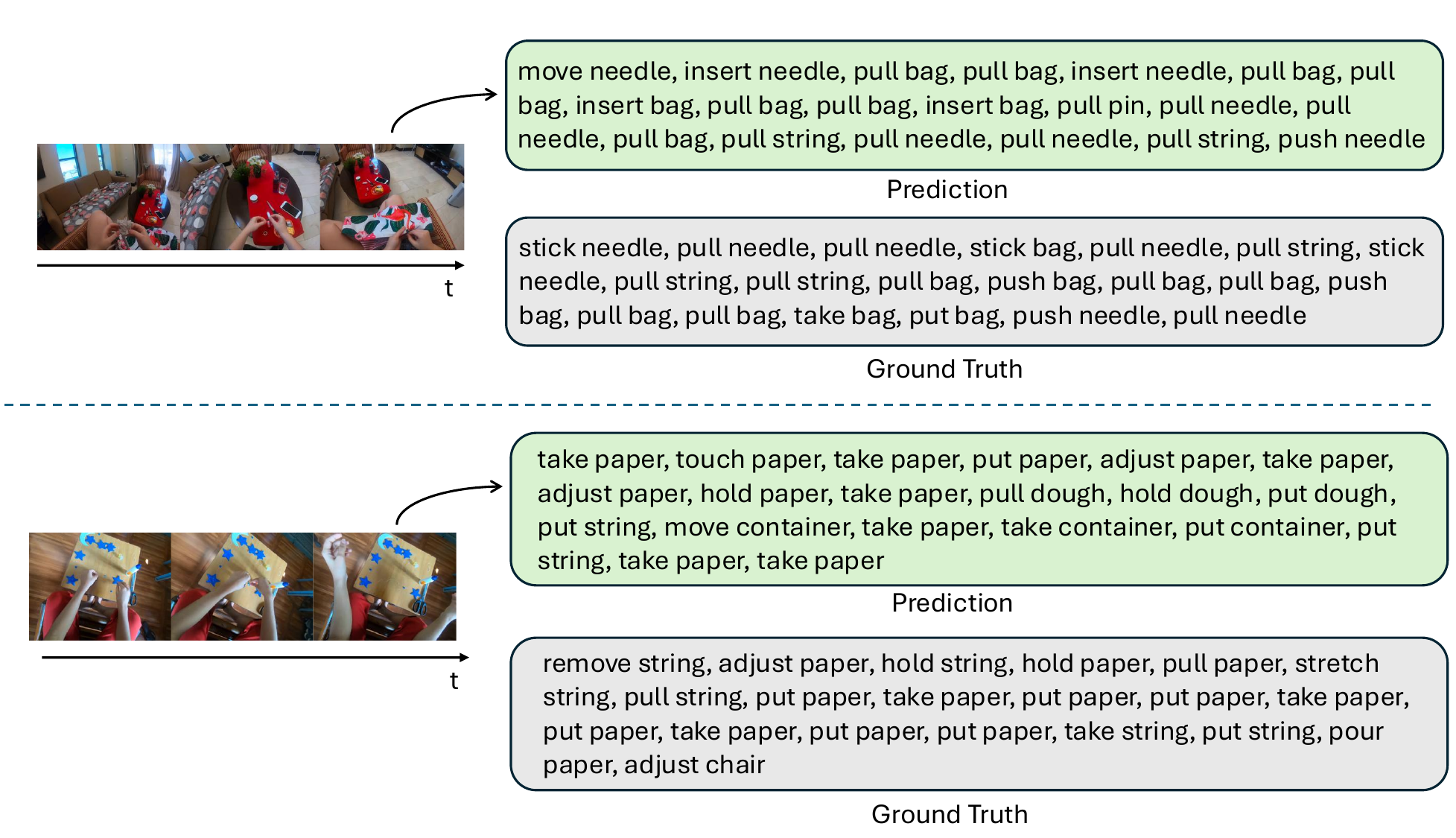}
    \caption{\textbf{Qualitative Results for Ego4D LTA.} Predicting long-term future actions are extremely challenging because the future is uncertain and there are multiple possible future action sequences. The action sequences produced by our method generally matches the person's goal and behavior. In the bottom subfigure, the model predicts ``dough'', ``container'' while the person is not in a cooking scenario. This suggests that perception ability of our model still has room for improvements.}
    \label{fig:vis_ego4d}
\end{figure*}

\noindent \textbf{LTA.}
Figure~\ref{fig:vis_ego4d} shows two examples from the Ego4D LTA task. From the figure we can see that our model is able to produce reasonable action sequences. Additionally, the model predicts ``dough'', ``container'' while the person is not doing cooking-related tasks. This indicates that our model's perception ability still has room for improvements. Finally, we can observe that verb prediction is more accurate than noun prediction. This shows the strong planning ability of our method.

% \newpage
% {
%     \small
%     \bibliographystyle{ieeenat_fullname}
%     \bibliography{main}
% }

\end{document}